%% file: main.tex
\documentclass[10pt,twocolumn,letterpaper]{article}

 \usepackage{cvpr}              


\definecolor{cvprblue}{rgb}{0.21,0.49,0.74}
\usepackage[pagebackref,breaklinks,colorlinks,allcolors=cvprblue]{hyperref}
\usepackage{booktabs}
\usepackage{multirow}

\def\paperID{11}
\def\confName{CVPR}
\def\confYear{2026}

\title{D2-V2X: Depth-Driven Cooperative V2X Reasoning for Autonomous Driving}

\author{
    Kevin Richard \quad Alphin Varghese \quad Colin Pham \quad David Oh \quad Srijan Das \\
    University of North Carolina at Charlotte \\
    {\tt\small \{krich103, avarghe3, cpham10, ioh1, sdas24\}@charlotte.edu}
}

\begin{document}
\maketitle
\input{sec/0_abstract}    
\input{sec/1_intro}
\input{sec/2_related_work}
\input{sec/3_method}
\input{sec/4_experiments}
\input{sec/5_conclusion}
{
    \small
    \bibliographystyle{ieeenat_fullname}
    \bibliography{main}
}

\end{document}

%% file: sec/0_abstract.tex
\begin{abstract}
Single-vehicle Vision-Language Models (VLMs) are fundamentally constrained by sensor occlusions. While Vehicle-to-Everything (V2X) systems mitigate this, current benchmarks lack the cooperative reasoning required for resolving ambiguities in complex environments. We introduce \textbf{D2-V2X}, a spatially-aware Question-Rationale-Answer (QRA) benchmark featuring 8,500 triplets derived from multimodal vehicle and infrastructure sensors. We additionally establish a baseline that aligns 3D LiDAR features with the VLM’s latent space. By enforcing natural language Chain-of-Thought rationales prior to structured JSON outputs, our model is forced to explicitly articulate spatial relations. Our experiments demonstrate that grounding VLMs in cooperative LiDAR achieves $24.4\%$ recall in identifying occluded hazards compared to near-zero in zero-shot models and reduces spatial estimation error for visible objects by $77\%$ compared to the zero-shot baseline. While the model achieves a functional decision-making F1-score of 53.5, we identify 3D-to-2D projection as a fundamental bottleneck in current VLM architectures, establishing a new baseline for future innovation. Data, code, and trained models available at \href{https://github.com/KevinRichard1/D2-V2X}{https://github.com/KevinRichard1/D2-V2X}
\end{abstract}

%% file: sec/1_intro.tex
\section{Introduction}
\label{sec:intro}
Modern Vision-Language Models (VLMs) have redefined scene understanding \cite{Cao_2024_CVPR}, yet their deployment in autonomous driving is fundamentally constrained by ego-centric sensing. In complex urban environments, limited detection ranges and sensor occlusions render single-vehicle perception insufficient for safe navigation.

While Vehicle-to-Everything (V2X) systems mitigate these constraints by integrating collaborative sensing with ego-vehicle perception, current VLM benchmarks still fail to address the cooperative spatial reasoning required to resolve 3D ambiguities. To address these challenges, we introduce D2-V2X, a grounded Question-Rationale-Answer (QRA) dataset. Unlike existing works \cite{qian2024nuscenes, ishihara2026stride, you2026v2x}, our QRA format forces the model to articulate a natural language rationale that explains the relations of occluded objects using multi-sensor data before committing to a final driving action. By grounding these rationales in multimodal features rather than 2D pixels alone, we ensure that the model’s reasoning is anchored in the physical geometry of the intersection.

\begin{figure}[t]
  \centering
  \includegraphics[width=\linewidth]{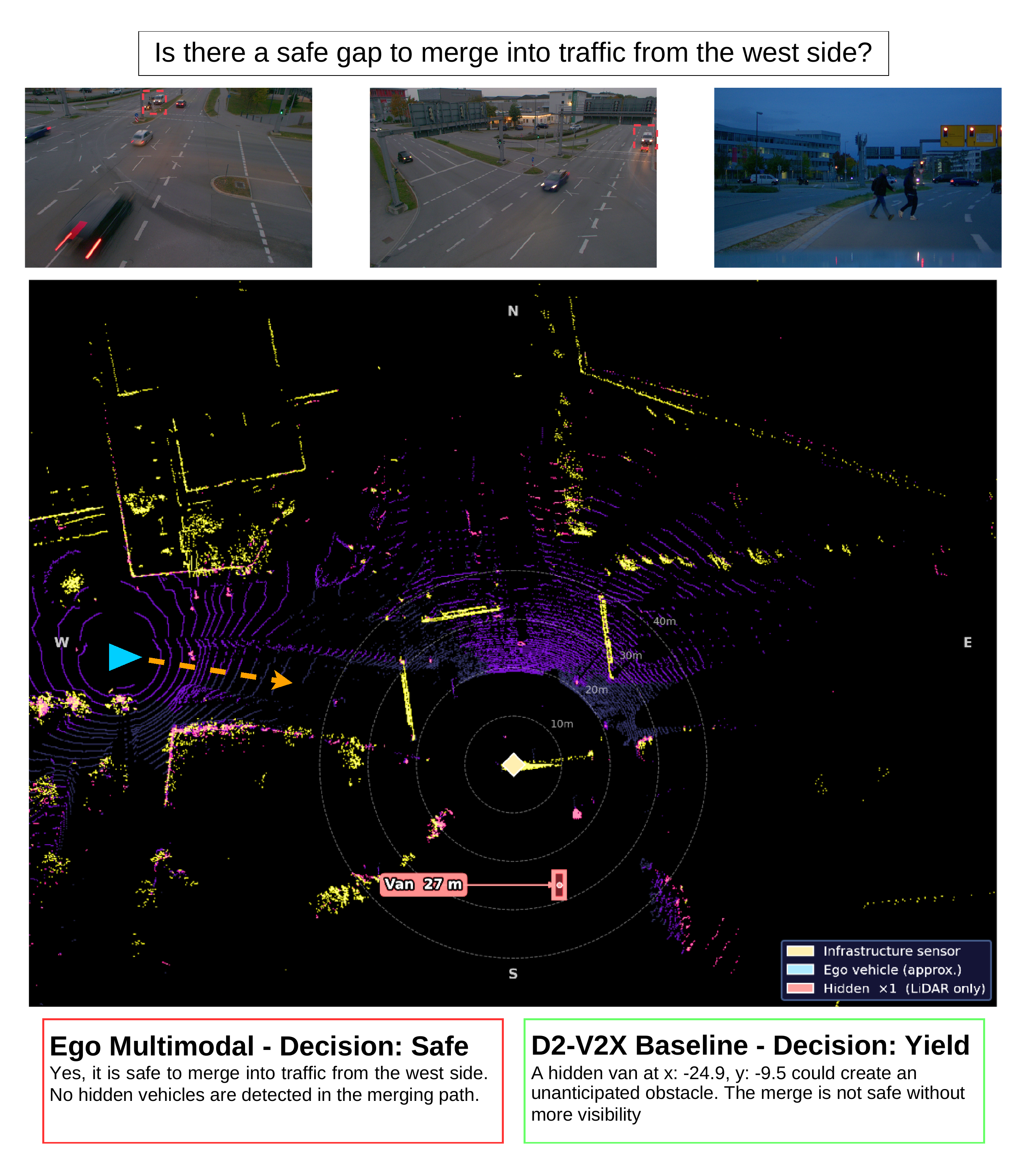}\vspace{-0.2in}
    \caption{\textbf{V2X Cooperative Perception.} While Ego-only multimodal models (left) fail to detect occluded hazards, our D2-V2X baseline (right) leverages infrastructure LiDAR to localize a hidden van (red box, 27m), enabling a safe decision in a complex scenario.} \vspace{-0.2in}
  \label{fig:teaser}
\end{figure}

Crucially, this addresses the absence of datasets and baselines that concurrently leverage 3D LiDAR, cooperative V2X infrastructure, and Chain-of-Thought (CoT) reasoning within a unified multimodal framework. D2-V2X establishes a foundation for VLMs to interpret and navigate complex, multi-sensor environments. To demonstrate the utility and difficulty of this benchmark, we introduce a parameter-efficient architecture that aligns early-fused LiDAR features with an ego-vehicle VLM. Through Supervised Fine-Tuning (SFT) on our 8,500 triplet dataset, we demonstrate that while articulating spatial relations significantly improves occlusion awareness and driving decision-making, the translation of these 3D features into 2D spatial coordinates remains a fundamental challenge.

Our contributions are summarized as follows:
\begin{itemize}
    \item We introduce \textbf{D2-V2X}, a novel multimodal benchmark designed to evaluate VLM reasoning using V2X sensors.
    \item We establish a baseline multimodal framework that improves occlusion detection recall to $24.4\%$ and reduces distance estimation error by $77\%$.
    \item We expose a critical architectural bottleneck in state-of-the-art VLMs: while they can accurately reason about 3D occlusions, they are ineffectual at 2D spatial projection, establishing an open challenge for future V2X models.
\end{itemize}

%% file: sec/2_related_work.tex
\begin{figure}[t]
    \centering
    \small
    \begin{tabular}{|p{0.95\linewidth}|}
    \hline
    \textbf{Task:} Spatial Awareness \\
    \hline
    \textbf{Q:} Checking for hidden vehicles in the southeast direction. Are there any? \\
    \textbf{R:} A van is visible, but a car at x: 25.04, y: -27.33 is obscured by the van, detected at 37.06 meters. This car could be a potential hazard if not monitored. \\
    \textbf{A:} Yes, a car is hidden by a van in the southeast direction at 37 meters. \newline
    \texttt{\{ "decision": "monitor", "hazard\_level": "medium", "count": 1,} \newline
    \texttt{\ \ "grounded\_objects": [\{ "type": "car",} \newline
    \texttt{\ \ \ "bbox": [720, 245, 862, 339], "distance\_m": 37.06,} \newline
    \texttt{\ \ \ "sensor\_id": "s110\_camera\_basler\allowbreak\_south1\_8mm" \}]\}} \\
    \hline
    \end{tabular}
    \caption{\textbf{Sample D2-V2X QRA triplet} The explicitly generated rationale forces the model to resolve 3D spatial geometry before outputting the structured JSON response.} \vspace{-0.2in}
    \label{fig:dataset_sample}
\end{figure}

\section{Related Work}
\label{sec:related_work}
\subsection{Autonomous Driving Models}
Recent works have demonstrated the potential of integrating language models for reasoned decision-making in autonomous driving \cite{park2024vlaad, sima2024drivelm}. Similarly, scene-understanding models have advanced spatial reasoning capabilities \cite{Cao_2024_CVPR}. Despite these advances, single-vehicle, image-only VLMs \cite{lubberstedt2025v3lma} are inherently constrained by the physical limits of a singular perspective. While the inclusion of Bird’s Eye View (BEV) maps moves towards spatial awareness \cite{brandstatter2025bev}, it lacks the 3D information for complete depth understanding.

Foundational perception works like V2X-ViT \cite{xu2022v2xvit} show that cooperative infrastructure and vehicle fusion significantly outperforms single-vehicle perception, particularly in highly occluded intersections. Attempts to bring language models into this space show improvements over ego-only perception, yet primarily focus on single-modality approaches \cite{wu2025v2x, chiu2025v2v}. By omitting LiDAR, they lack the precise 3D spatial grounding required for accurate distance estimation and robust hazard detection in complex scenarios.

\subsection{Driving QA Datasets}
Existing structured QA benchmarks rely on single-vehicle datasets, limiting their use for testing cooperative perception \cite {qian2024nuscenes, ishihara2026stride}. Recent V2X benchmarks \cite{zhou2025tumtraf, you2026v2x} provide multi-perspective data but evaluate via multiple-choice or end-to-end tasks, lacking 3D-grounded explainability. D2-V2X addresses this gap with a cooperative, multimodal benchmark utilizing Chain-of-Thought rationales to explicitly ground decisions in spatial data.

\begin{figure*}[t]
    \centering
    \includegraphics[width=\textwidth]{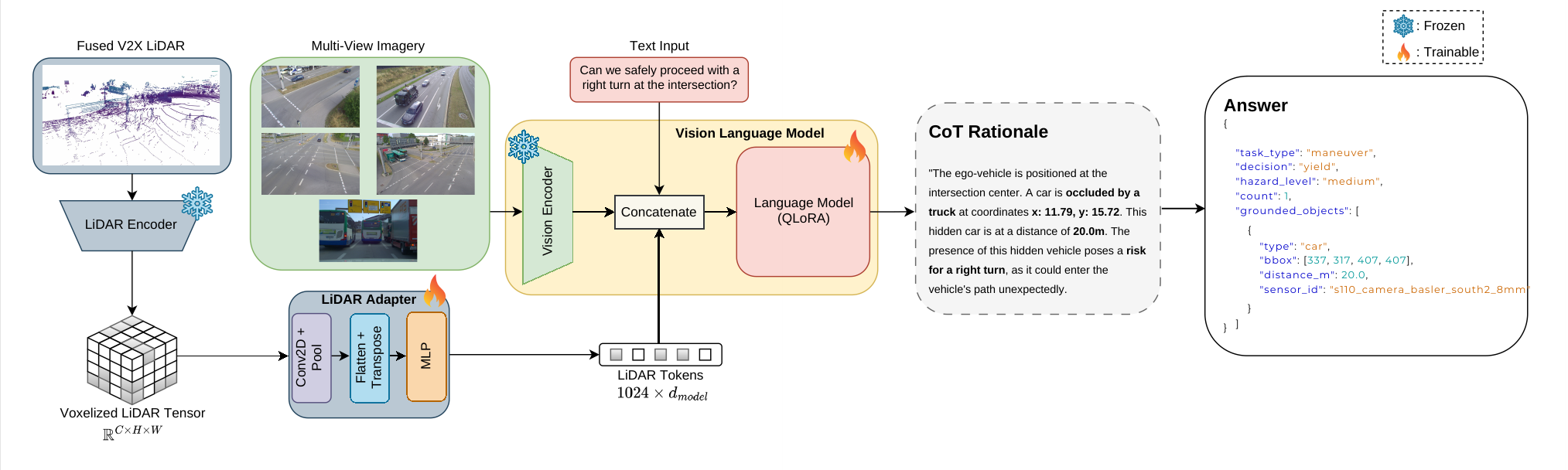}
    \caption{\textbf{D2-V2X Architecture.} A pre-trained CenterPoint backbone extracts 3D features from fused V2X LiDAR, which are projected into 1D tokens. These are concatenated with multi-view image and text embeddings. The QLoRA-tuned Qwen3-VL processes this sequence to generate grounded reasoning and a structured decision.}
    \label{fig:architecture}
\end{figure*}

%% file: sec/3_method.tex
\section{Method}
\subsection{Dataset Generation}
We build the D2-V2X dataset upon TUMTraf-V2X \cite{zimmer2024tumtraf}, utilizing four infrastructure cameras, the ego-camera, and early-fused V2X LiDAR. Lacking public test set ground truth, we apply a scene-stratified split to the original train set, yielding a new distribution: train ($76\%$), val ($13\%$), and test ($11\%$) set. 

To define ground truth for occluded objects, we apply a geometric heuristic to the cooperative LiDAR. We project the angular boundaries of 3D bounding boxes relative to the ego-origin into a 2D Bird’s-Eye View (BEV) plane, sorted by depth. Objects whose angular profiles are heavily intersected by closer obstacles are labeled as occluded, rendering it invisible to the ego-camera while remaining localized in the fused LiDAR.

This spatial metadata prompts GPT-4o \cite{hurst2024gpt} to generate our Question-Rationale-Answer (QRA) triplets across three types of tasks: Spatial Awareness ($30\%$), Scene Counting ($30\%$), and Driving Maneuvers ($40\%$). These tasks evaluate relative object localization (Spatial), occluded object frequency (Counting), and action-planning safety (Maneuvers). Additionally, triplets are generated using four distinct personas to increase diversity in the dataset. We provide the model with the scene representations, including both raw metrics such as 3D coordinates, distances, and bounding boxes, as well as natural language descriptors such as position and heading.

To ensure data integrity, an automated validation pipeline cross-references all generated triplets against the original ground truth. This pipeline enforces internal consistency (e.g., matching object counts) and filters hallucinations by verifying that generated object distances fall within a dynamic tolerance range. The initially generated 9,000 samples were passed through this pipeline, and approximately 500 were filtered out due to object hallucinations or errors in spatial consistency from the ground truth input to the generated responses. Additional human-in-the-loop evaluation was conducted on a random $1\%$ subset of the data to maintain semantic and stylistic consistency, with all evaluated samples passing without requiring modification. Through this process, we obtain a total of 8,500 QRA triplets.

\subsection{D2-V2X Benchmark}
We establish a baseline by adapting the Qwen3-VL-8B-Thinking model \cite{bai2025qwen3} using SFT. We prepare the multimodal training data by extracting 3D spatial features from early-fused ego-vehicle and infrastructure LiDAR point clouds provided in the TUMTraf dataset \cite{zimmer2024tumtraf}. Instead of using flattened Bird's Eye View (BEV) maps, 3D spatial features are extracted using a frozen, pre-trained CenterPoint backbone \cite{yin2021center} and saved as intermediate voxelized tensors. With these inputs, the model is trained to produce a natural language rationale grounded in sensor data, followed by a formatted JSON object. 

To bridge the modality gap, we design a lightweight projection layer. The pre-trained CenterPoint backbone outputs a dense spatial feature map $\mathbf{V} \in \mathbb{R}^{C \times H \times W}$. The adapter $f(\mathbf{V})$ uses a 2-layer 2D convolutional stem and MLP to project $\mathbf{V}$ into a sequence of $N=1024$ tokens within the model's latent space $d_{model}$. Formally, for a given input query, the final multimodal token sequence $\mathbf{E}_{input}$ is constructed by sequentially concatenating the image embeddings $\mathbf{E}_{img}$, the projected LiDAR tokens, and the text embeddings $\mathbf{E}_{txt}$:
$$\mathbf{E}_{input} = [ \mathbf{E}_{img} \parallel f(\mathbf{V})^\top ) \parallel \mathbf{E}_{txt} ]$$
where $\parallel$ denotes sequence concatenation along the token dimension, and $\mathbf{E}_{input} \in \mathbb{R}^{(L_{img} + 1024 + L_{txt}) \times d_{model}}$. This ordering ensures dimensional alignment and preserves the positional integrity expected by the VLM during both training and inference.

\begin{table}[t]
\centering
\footnotesize
\resizebox{\columnwidth}{!}{%
\begin{tabular}{@{}lcccccc@{}}
\toprule
\textbf{Method} & \textbf{F1} & \textbf{Occ.} & \textbf{Occ.@10m} & \textbf{Vis. MAE} & \textbf{BERT} & \textbf{mIoU} \\\midrule
Qwen3-VL (Zero-Shot) & 0.22 & 0.00 & 0.00 & 40.34 & 0.69 & 0.00 \\
Image w/ BEV (SFT) & \textbf{0.54} & 0.14 & 0.07 & 8.98 & \textbf{0.85} & \textbf{0.06} \\
Ego Multimodal (SFT) & 0.45 & 0.16 & 0.08 & 8.83 & \textbf{0.85} & 0.01 \\
D2-V2X w/o Rationale & 0.48 & 0.03 & 0.02 & \textbf{7.58} & \textbf{0.85} & 0.01 \\ \midrule
\textbf{D2-V2X (Full)} & \textbf{0.54} & \textbf{0.24} & \textbf{0.11} & 9.16 & 0.84 & 0.01 \\ \bottomrule
\end{tabular}
}
\caption{\textbf{Aggregate Performance.} D2-V2X improves decision-making (F1), occlusion recall (Occ.), and precise localization (Occ.@10m). Stable BERTScore confirms rationale quality, while mIoU and Vis. MAE highlight the spatial projection bottleneck.\label{tab:aggregate_results}}
\vspace{-0.2in}
\end{table}

\begin{table*}[t]
\centering
\footnotesize
\setlength{\tabcolsep}{4pt}
\begin{tabular}{@{}l|cccc|cccc|cccc@{}}
\toprule
\multirow{3}{*}{\textbf{Method}} & \multicolumn{4}{c|}{\textbf{Spatial (Grounding)}} & \multicolumn{4}{c|}{\textbf{Counting (Obstacles)}} & \multicolumn{4}{c}{\textbf{Maneuver (Planning)}} \\ \cmidrule(l){2-13} 
 & F1 & Occ. & Occ.@20m & Vis. MAE & F1 & Occ. & Occ.@20m & Vis. MAE & F1 & Occ. & Occ.@20m & Vis. MAE \\
 \midrule
Qwen3-VL (Zero-Shot) & 0.04 & 0.00 & 0.00 & 38.09 & 0.02 & 0.00 & 0.00 & 42.80 & 0.27 & 0.00 & 0.00 & 33.96 \\
Image w/ BEV (SFT) & 0.33 & 0.24 & 0.11 & \textbf{9.26} & 0.25 & 0.08 & 0.03 & 9.00 & \textbf{0.43} & \textbf{0.19} & \textbf{0.15} & 8.47 \\
Ego Multimodal (SFT) & 0.32 & \textbf{0.35} & \textbf{0.17} & 9.92 & 0.30 & 0.13 & 0.08 & 8.34 & 0.32 & 0.12 & 0.11 & 9.61 \\
D2-V2X w/o Rationale & 0.25 & 0.09 & 0.02 & 10.30 & 0.25 & 0.04 & 0.04 & \textbf{6.34} & 0.37 & 0.00 & 0.00 & \textbf{6.78} \\ \midrule
\textbf{D2-V2X (Full)} & \textbf{0.38} & 0.33 & 0.12 & 10.69 & \textbf{0.33} & \textbf{0.29} & \textbf{0.13} & 8.12 & 0.40 & 0.15 & 0.13 & 10.17 \\ \bottomrule
\end{tabular}
\caption{\textbf{D2-V2X Task-Specific Performance.} While cooperative V2X features significantly improve occlusion recall in global reasoning tasks like Scene Counting, the higher-fidelity Ego Multimodal baseline retains a slight edge in highly localized Spatial tasks. (Note: Rationale and mIoU metrics are consistent across tasks and reported in Table 1).\label{tab:task_breakdown}}
\end{table*}

To enable the model to understand the new modality while maintaining computational efficiency, we employ a two stage training strategy using 4-bit Quantized Low-Rank Adaptation (QLoRA) \cite{dettmers2023qlora}. During the initial warmup stage, the VLM’s weights are frozen, and the adapter is trained for one epoch on the QRA dataset with a learning rate of $1 \times 10^{-3}$ to prevent the randomly initialized projector from corrupting the pre-trained model representations. In the second stage, we utilize a QLoRA configuration with a rank $r=64$ and $\alpha=128$, targeting all linear layers in addition to the unfrozen adapter in order to capture the complexities of the multimodal data. The model is instruction-tuned for three epochs with a learning rate of $2 \times 10^{-5}$. Throughout the training, we use the AdamW optimizer with a weight decay of $0.05$. All experiments are conducted on a single NVIDIA A100 GPU using an effective batch size of 64.

%% file: sec/4_experiments.tex
\section{Experiments}
We evaluate our D2-V2X baseline against the zero-shot Qwen3-VL-8B-Thinking model \cite{bai2025qwen3} and three fine-tuned variants: an ego-only model restricted to single-vehicle sensors, an image-based model utilizing V2X imagery in addition to LiDAR BEV projections, and a direct-answer ablation that omits the intermediate reasoning step.

\paragraph{Evaluation Metrics} Occlusion Recall (Occ.) measures the percentage of hidden objects successfully matched to a model prediction. Occ.@10m and Occ.@20m, calculate recall strictly within a 10m or 20m distance threshold. We use 10m for aggregate reporting (Table 1) to evaluate high-precision localization, and 20m for task-specific breakdowns (Table 2) to accommodate variance in distant Scene Counting tasks. Vis. MAE represents distance estimation error for correctly matched visible objects. Decision F1 is macro-averaged across four action classes in Table 1 and locally within specific sub-tasks in Table 2.

\subsection{Performance Comparison}
Cooperative infrastructure data significantly improves global scene understanding, increasing in decision-making F1 by $20\%$ and occlusion recall by $50\%$ over the single-vehicle model. However, the simpler Image w/ BEV model achieves better bounding box projection (mIoU) and lower distance estimation errors (MAE). Generating a natural language rationale introduces a compromise between spatial precision and action-planning. Training the multimodal model without rationales improves distance MAE by $17.2\%$ but degrades occlusion recall by $87.5\%$ While D2-V2X improves occlusion awareness, MAE remains too high for safe, real-world deployment. This distance accuracy regression is likely a result of our specific CoT formulation, in which the model prioritizes hidden hazards and maneuver safety. While high SFT BERTScores \cite{Zhang*2020BERTScore:} confirm stable linguistic formatting, 3D-to-2D spatial projection remains a challenge. Despite identifying objects and their distances, the model struggles with converting detected 3D objects into a 2D bounding box in the ego-vehicle’s image plane.

\subsection{Task-Specific Performance}
We employ a consistent suite of metrics across the Spatial Awareness, Scene Counting, and Driving Maneuvers tasks to evaluate the model’s overall spatial reasoning. Interestingly, while V2X fusion benefits global scene understanding, we observe an inverse effect on highly localized tasks. Because the fused V2X LiDAR provides a more complete global context in spatial tasks, downsampling is significantly more aggressive. This results in a loss of local information compared to the ego-only model, which benefits from a higher-fidelity representation of the immediate 3D scene. In Driving Maneuvers, the Image w/ BEV baseline outperforms the full D2-V2X model in F1 score. We attribute this to the top-down BEV image providing a much stronger, direct spatial representation for simple pathing decisions, whereas the full model struggles to extract navigational geometry from highly compressed 3D V2X tokens. However, the benefits of cooperative reasoning are evident in counting tasks, where the occlusion recall at a threshold of $20$m increases by $62.5\%$ over the single-vehicle model and more than triples compared to the direct response model. Lastly, the $8\%$ relative gain in Maneuver F1 over the non-rationale ablation suggests that while V2X LiDAR provides the necessary raw data, the chain-of-thought rationale is critical to correctly interpret those features for successful autonomous navigation.

%% file: sec/5_conclusion.tex
\section{Conclusion}
We propose the \textbf{D2-V2X} benchmark, a spatially-aware Question-Rationale-Answer (QRA) dataset of 8,500 triplets. Our results demonstrate that incorporating voxelized cooperative LiDAR features provides crucial global context and depth to the VLM’s spatial reasoning. By including natural language rationales, the model successfully bridges spatial-precision and action-planning. However, we highlight several limitations. First, the dataset derives from a single intersection, and relying on an LLM for annotation introduces a risk of unverified biases. Second, our 2D geometric occlusion heuristic ignores 3D height profiles where taller vehicles may remain partially visible. Third, our baseline assumes ideal data transmission, neglecting real-world challenges like communication latency. Architecturally, the compression of dense LiDAR scenes into a static sequence of 1024 tokens and the reliance on early fusion without cross-attention, limits fine-grained multimodal alignment, contributing to the 3D-to-2D spatial projection bottleneck. Future work will focus on scaling to diverse intersections, incorporating realistic network noise, and supporting higher-resolution 3D tokenization to advance cooperative autonomous driving